\documentclass[letterpaper, 10 pt, conference]{ieeeconf}  

\usepackage{amssymb}
\usepackage{amsmath}
\usepackage{commath}
\usepackage{mathtools}
\DeclarePairedDelimiter{\ceil}{\lceil}{\rceil}
\usepackage{mathbbol}
\usepackage{hyperref}
\usepackage{import}
\usepackage{paralist}
\usepackage[pdf]{svg}
\usepackage{gensymb}
\DeclareSymbolFontAlphabet{\amsmathbb}{AMSb}

\IEEEoverridecommandlockouts                              

\usepackage{graphics} 
\usepackage{epsfig} 
\usepackage{todonotes}
\usepackage[normalem]{ulem}

\title{\LARGE \bf
A Predictive Momentum-Based Whole-Body Torque Controller:\\ 
Theory and Simulations for the iCub Stepping
}

\author{Stefano Dafarra, Francesco Romano, Gabriele Nava, Francesco Nori
\thanks{This work was supported by the FP7 EU project CoDyCo (No. 600716 ICT 2011.2.1 Cognitive Systems and Robotics) and Horizon 2020 EU project An.Dy. (No. 731540 Research and Innovation Programme)}%
\thanks{The authors are with the iCub Facility Department, Istituto Italiano di Tecnologia, 16163 Genova,
Italy (e-mail: name.surname@iit.it)}}

\DeclareMathOperator*{\minimize}{minimize}

\begin{document}
	
\maketitle
\thispagestyle{empty}
\pagestyle{empty}

\begin{abstract}
	When balancing, a humanoid robot can be easily subjected to unexpected disturbances like external pushes. In these circumstances, reactive movements as steps become a necessary requirement in order to avoid potentially harmful falling states. 
	In this paper we conceive a Model Predictive Controller which determines a desired set of contact wrenches by predicting the future evolution of the robot, while taking into account constraints switching in case of steps. The control inputs computed by this strategy, namely the desired contact wrenches, are directly obtained on the robot through a modification of the  momentum-based whole-body torque controller currently implemented on iCub.
	The proposed approach is validated through simulations in a stepping scenario, revealing high robustness and reliability when executing a recovery strategy.
\end{abstract}

\section{Introduction}\label{c6sec:soa}
The unpredictable nature of the real world is one of the leading obstacles to the widespread diffusion of robots outside labs. These complicated mechanical systems ought to be robust against a wide spectrum of disturbances. Considering humanoid robots, the aim for robustness translates into preventing stumbling. Despite the apparent straightforwardness of such an objective, complications arise when facing it from an algorithmic point of view.    
Considering humanoid robots, problems arise due to their intrinsic underactuation \cite{Spong1998} which, in essence, means that by exploiting all theirs actuated degrees-of-freedom they can control their internal configuration, but they cannot affect directly their global pose.
For example, an astronaut in the space is free to move all his limbs, but without exploiting any hooking he cannot move inside the spaceship.
On the contrary, by exploiting contacts with the environment it is possible to circumvent this limitation.
Additional difficulties arise by the fact that contacts may change over time. This results in a different evolution of the constrained dynamical system making the overall system hybrid \cite{lygeros1999hybrid}, i.e. it possesses both a continuous and  discrete time dynamics.

An appealing research problem consists in applying Model Predictive Control (MPC) techniques to these particular systems. MPC is a particular optimal control method which enables the introduction of the feedback into the optimization procedure thanks to the ``Receding Horizon Principle'' \cite{Mayne90MPC,de2000stability,Mayne2000Stability}. 
The basic concept consists in solving, at each time step, a new finite-horizon optimal control problem initialized at the current plant state.
At every time instant only the first control input is then applied to the plant.
Model predictive controllers are appealing for controlling hybrid systems \cite{lazar2006stabilizing,bemporad2002hybrid} since the full hybrid model can be exploited. 
Indeed, thanks to the prediction capabilities, it is possible to include inside the formulation both time- and state-dependent switching, performing anticipatory actions for the imminent variation in the dynamics. However MPC does not solve those problems related to the numerical integration of hybrid systems, which indeed is an open research problem.

In literature the original underactuated hybrid dynamics is usually scaled down to simplified models, making the problem easier to be handled.  Simple models like the Linear Inverted Pendulum (LIP) \cite{Kajita2001} are widely adopted. In this context, MPC has been applied in order to stabilize walking patterns \cite{wieber2006trajectory,diedam2008online,missura2014balanced}, especially of position controlled robots. The result is usually a slow and steady walking style characterized by a nearly constant Center of Mass (CoM) height. The LIP model has been applied also in \cite{stephens2010pushforce} with a robot equipped with force controlled joints. While being easily applicable, this simple model provides only a limited amount of informations about the actual dynamics of the robot. A related popular approach is based on the Capture Point framework \cite{Pratt2006}. While keeping the model complexity low, this technique allows to evaluate the possibility of the robot to stay in the upright position. Starting from the simple Linear Inverted Pendulum, the model has been progressively enriched \cite{pratt2012capturability} to catch different robot peculiarities. This method is particularly interesting due to the possibility of drawing stability criteria \cite{englsberger2011bipedal} and in \cite{krause2012stabilization} authors applied MPC techniques based on the Capture Point formulation.
A recent trend seems to move in a different direction. The CoM dynamic model is substituting the simple pendulum, especially when planning complex trajectories. In \cite{dai2014whole,herzog2015trajectory,carpentier2016versatile} a kinematic planner is merged with one based on the CoM dynamics to accomplish complex and feasible motions, while taking into account different contact locations. The main issues of this approach are usually related to the computational complexity, while the optimization problem is generally non-convex, introducing problems of local minima. In other applications, only the momentum dynamics is considered. In \cite{caron2016multi} only the 3D CoM acceleration is taken into consideration, while in \cite{daiplanning} authors propose a convex upper bound of the angular momentum to be minimized. In \cite{ponton2016convex} the derivative of the angular momentum is approximated by using quadratic constraints together with slack variables necessary to keep the approximation error low. Nevertheless, in this approach, it is not possible to directly penalize the use of the angular momentum, while introducing many additional variables into the formulation.

Following this direction, we present here a momentum-based whole-body torque controller based on a MPC formulation. 
In particular, the dynamic evolutions of the robot linear and angular momentum are taken into account. We thus make use of a \emph{reduced} model: differently from simplified models used in literature, the robot momentum is an exact model that captures the \emph{global} behaviour of the robot. We deal with the complications introduced by the derivative of the angular momentum by resorting to a Taylor expansion. The presented controller allows to deal directly with the intrinsic hybrid dynamic of the system by considering time-varying constraints. 
The peculiarity of our approach also resides in the fact that the computed control inputs, i.e. the contact wrenches, are directly applied on the robot, rather than used to define a joint position reference trajectory. In particular, the presented scheme inherits its structure from the momentum-based whole body torque controller \cite{Frontiers2015,pucci2016highly,nava16} implemented on iCub. Torque control is particularly suitable for our application given that it permits to absorb the impacts efficiently, maintaining the balance also in case of robot positioning errors.
We tested this approach on the iCub humanoid robot while performing a step recovery strategy.


\section{Background} \label{sec:background}
\subsection{Notation}
Throughout this paper we adopt the following notation. 
\begin{itemize}
	\item $\mathcal{I}$ represents an inertial reference frame with the origin placed on the ground, while the orientation is so that the $z-$axis points against the gravity and the $x-$axis is oriented frontally with respect to the robot. 
	\item ${1}_n$ represents a $n \times n$ identity matrix. $0_{n \times m} \in \amsmathbb{R}^{n\times m}$ is a zero matrix while $0_n = 0_{n \times 1}$ is a zero column vector of size $n$.
	\item $x_\text{CoM} \in  \amsmathbb{R}^3$ is the position of the center of mass with respect to $\mathcal{I}$.
	\item $\norm{x}^2_W = x^\top W x$ is the weighted square norm of $x$.
	\item $x^\wedge \in \amsmathbb{R}^{3 \times 3}$ denotes the skew-symmetric matrix such that $x^\wedge y = x \times y$, where $\times$ denotes the cross product operator in $\amsmathbb{R}^3$.
	\item $\ceil{\cdot}$ denotes the ceiling operator which outputs the rounding of the argument toward $+\infty$.  
\end{itemize}

\subsection{Whole-body torque control}
\label{sec:momentum}

The predictive controller we propose in this paper, relies on a whole-body optimization-based torque controller for the stabilization of the desired contact wrenches.
The original whole-body control algorithm is a momentum-based hierarchical controller composed of two control objectives \cite{pucci2016highly,nava16}.
The first, and most priority objective, is the tracking of a desired robot momentum while the second is the stabilization of the zero dynamics.

In this paper we slightly modify the original controller by directly commanding desired contact wrenches.
We thus ``substitute'' the first control objective, i.e. generation of contact wrenches to track a desired robot momentum, with our controller proposed in this paper.

The second objective, which is left as in the original controller, is responsible for constraining the joint variables and avoid internal divergent behaviours \cite{nava16}. When described as an optimization problem, this second objective can be formulated as:
\begin{subequations}
    \label{eq:zero_stab_min}
\begin{align}
            \min_\tau & \quad \norm{\tau - \psi}^2  \label{eq:zero_stab_min_cost}\\
            \text{s.t.}&\quad M(q)\dot{\nu} + h(q, \nu) - J^\top f^{ref} = B \tau   \label{eq:zero_stab_min_dyn}\\
                       &\quad J \dot{\nu} + \dot{J} \nu = 0    \label{eq:zero_stab_min_constr}\\
                       &\quad \psi := h_j(q, 0) - J^{(j),\top} f^{ref}  - K_p^j(q_j - q_j^{ref}) - K_d^j \dot{q}_j  \label{eq:zero_stab_min_post}
\end{align}
\end{subequations}
Eq.\eqref{eq:zero_stab_min_dyn} describes the free-floating dynamics of the mechanical system \cite[Ch. 7]{Siciliano2009}, where $(q, \nu)$ is the free-floating state of the robot.
Eq.\eqref{eq:zero_stab_min_constr} is the constraint equation describing the kinematic constraints associated with the contacts.
Eq.\eqref{eq:zero_stab_min_post}, which resembles a PD plus gravity and contact wrenches compensation, plays the role of a desired joint torque reference where $h_j$ and  $J^{(j)}$ denotes the joint space bias term and Jacobian respectively.
Note that the controller depends on two inputs: $f^{ref}$ and $q_j^{ref}$.
As we will show later, our predictive controller is responsible of choosing the former quantity, while the latter is the output of a inverse kinematics algorithm.


%

\section{Problem Formulation} \label{sec:problem_formulation}
\subsection{The model} \label{sec:model}
In this work, we examine the scenario where the only contacts available are those expressed at the feet. 
Without loss of generality, we consider as a test case the robot balancing on single support and performing a step adopting the right foot as swing limb. 
The rate of change of the robot momentum, when expressed in a frame located at the center of mass with the same orientation of $\mathcal{I}$,
has the following expression:
\begin{equation}\label{c6eq:model_2f}
\begin{bmatrix}
\ddot{x}_\text{CoM}\\
\dot{H}_\text{ang}
\end{bmatrix} {=} \begin{bmatrix}
				m^{-1}{1}_3 & 0_{3\times3}\\
				0_{3\times3} & {1}_3
				\end{bmatrix}\left[
				{}^{\text{CoM}}X_{l} \: {}^{\text{CoM}}X_{r}
				\right]\begin{bmatrix}
									f_l\\f_r
							\end{bmatrix} {+} \bar{g}
\end{equation}
where $m \in \amsmathbb{R}$ is the mass of the robot and $\bar{g}$ the 6D gravity acceleration. The indexes $l$ and $r$ refer to left and right foot respectively. $f_i \in \amsmathbb{R}^6$, with $i$ either $l$ or $r$, is the contact wrench, composed of 3D forces and torques applied at the contact point, namely $f_i=\left[ \mathbb{f}_i^\top, ~ \tau_i^\top \right]$.  Matrices ${}^{\text{CoM}}X_{i}$, have the following particular structure:
\begin{equation}
{}^{\text{CoM}}X_{i} = \begin{bmatrix}
{1}_3 & 0_{3\times 3}\\
(x_{i}-x_{\text{CoM}})^{\wedge} & {1}_3
\end{bmatrix}.
\end{equation} 
The presence of $x_i$, highlights the dependency of the momentum from the foot position, but the presented model does not carry any information about the robot kinematics. The variables involved in the formulation do not affect directly $x_i$, thus it is not possible to define its feasible values properly. In addition, due to the very dynamic task the robot is going to perform, the tracking of a reference for the swing foot may not be perfect. Consequently, local modifications of a preplanned foot trajectory might not have a relevant role. 

In literature the foot positioning problem is usually addressed by exploiting the LIP model, as in \cite{feng2016robust}, \cite{ramos2014whole} or \cite{herdt2010online}. The kinematics limitations can be introduced through simple box constraints, which may not have a direct connection with the actual robot kinematic limits. As a consequence we rely on external planners to decide where to place the foot. This information will be considered as a datum for the presented strategy, thus keeping its computational complexity low.

For similar reasons we can assume to know the instant where the impact is going to happen, here called $t_{impact}$. Given the application we consider, the time necessary to take the step should be as little as possible. Consequently, it has been chosen to be the minimum time the robot needs to physically perform the step given its limits. This happened to be a common assumption in literature (\cite{feng2016robust, herdt2010online, herzog2015trajectory}), since it also avoids to introduce integer variables into the optimal control formulation, thus avoiding \textit{np-hard} complexity.

\subsection{The angular momentum}\label{sec:ang_mom}
The second set of rows of Eq.\eqref{c6eq:model_2f}, represents the derivative of angular momentum $\dot{H}_{\text{ang}}$, i.e.:
\begin{equation}
\dot{H}_\text{ang} = \sum_{i}^{l,r} (x_{i}-x_{\text{CoM}})^\wedge \: \mathbb{f}_i+\tau_i .
\end{equation}
If we consider both $x_\text{CoM}$ and $\mathbb{f}_i$ as two optimization variables, its product renders the formulation non-convex and thus much harder to be solved, mainly due to possible local minima. 
In our application, we can admit a certain level of approximation, since during the step this quantity does not need to be precisely controlled to zero. On the other hand, angular momentum can be used to penalize the usage of contact wrenches. For that purpose the Taylor expansion (around the last available values of $\mathbb{f}_i$ and $x_{\text{CoM}}$) truncated to the first order represents an acceptable approximation: 
\begin{IEEEeqnarray*}{RCL}
\IEEEyesnumber \phantomsection
\dot{H}_\text{ang} &\approx& \sum_{i}^{\{l,r\}} \tau_i + \left(x_{i}-x_{\text{CoM}}^0\right)^\wedge \: \mathbb{f}_i^0+ \nonumber\\
&+& \left(x_{i}-x_{\text{CoM}}^0\right)^\wedge \: \left(\mathbb{f}_i-\mathbb{f}_i^0\right) +  \nonumber \\
&+& \left(\mathbb{f}_i^0\right)^\wedge \: \left(x_{\text{CoM}}-x_{\text{CoM}}^0\right)   \nonumber\\
\Rightarrow \dot{H}_\text{ang} &\approx& \sum_{i}^{\{l,r\}} \tau_i + \left(x_{i}-x_{\text{CoM}}^0\right)^\wedge \: \mathbb{f}_i +  \\
&+& \left(\mathbb{f}_i^0\right)^\wedge \: \left(x_{\text{CoM}}-x_{\text{CoM}}^0\right).\IEEEyesnumber 
\end{IEEEeqnarray*}
Notice that the anticommutative property of the cross product, i.e.  $x^\wedge y = -y^\wedge x$, has been exploited. The superscript ${}^0$ refers to the last feedback retrieved from the robot, which will be used as the point around which we compute the Taylor expansion. By truncating at the first order, the approximation error is  $o\left(\left(\mathbb{f}_i-\mathbb{f}_i^0\right)\left(x_{\text{CoM}}-x_{\text{CoM}}^0\right) \right)$.
Thus, the less the CoM moves from the feedback position the better the approximation will be. This is usually the case given short prediction windows.

Finally, we define $\gamma$ as the state and $f$ as the control variable for the MPC problem, 
\begin{IEEEeqnarray}{rCl}
\IEEEyesnumber \phantomsection \label{c6eq:gamma}
\gamma &:=& \begin{bmatrix}
x_\text{CoM}^\top&
\dot{x}_\text{CoM}^\top&
H_\text{ang}^\top
\end{bmatrix}^\top \in \amsmathbb{R}^9,\IEEEyessubnumber\\
f &:=& \begin{bmatrix}
f_l^\top&
f_r^\top
\end{bmatrix}^\top \in \amsmathbb{R}^6 \IEEEyessubnumber
\end{IEEEeqnarray}
then, we can rewrite Eq.\eqref{c6eq:model_2f} as:
\begin{equation}\label{c6eq:model}
\dot{\gamma} = \tilde{E}v_\gamma \gamma + \tilde{F}_\gamma f + \tilde{G}_\gamma + \tilde{S}_\gamma^0
\end{equation}
where
\begin{IEEEeqnarray*}{RCL}
\tilde{E}v_\gamma &=& \begin{bmatrix}
0_{3\times3} & {1}_3 & 0_{3\times3}\\
0_{3\times3} & 0_{3\times3}  & 0_{3\times3}\\
\left(\mathbb{f}_l^0+\mathbb{f}_r^0\right)^\wedge & 0_{3\times3}  & 0_{3\times3}\\
\end{bmatrix}, \\
\tilde{F}_\gamma &=& \begin{bmatrix}
	                        0_{3\times 3} &0_{3\times 3} &0_{3\times 3} &0_{3\times 3} \\
				           m^{-1}1_3 & 0_{3\times 3} & m^{-1}1_3 & 0_{3\times 3}\\
				           \left(x_{l}-x_{\text{CoM}}^0\right)^\wedge & 1_3 & \left(x_{r}-x_{\text{CoM}}^0\right)^\wedge & 1_3					           		
			        \end{bmatrix},\\
\tilde{G}_\gamma &=& \begin{bmatrix}
0_5\\
-g\\
0_3
\end{bmatrix},\quad \tilde{S}^0_\gamma = \begin{bmatrix}
							  0_6\\
							  -\left(\mathbb{f}_l^0+\mathbb{f}_r^0\right)^\wedge x_{\text{CoM}}^0
                       \end{bmatrix}.
\end{IEEEeqnarray*}
Here, $\tilde{S}^0_\gamma$ introduces the constant terms resulting from the Taylor approximation.
Under the above considerations, the model described by Eq.\eqref{c6eq:model} is affine and can be easily inserted into a QP formulation, as described in Sec. \ref{sec:qp}.

\subsection{Constraints definition}\label{c6sec:constraints}

The constraints introduced in the formulation are mainly physical limitations induced by contacts. Considering contacts located at the feet, those constraints enforce the feasibility of the exerted wrenches. For example, the applied wrenches must be within the friction cone (approximated through a polyhedral convex cone) while their point of application (i.e. the center of pressure, CoP) should lie inside the support polygon.
Formally, we can write these constraints linearly as follows:
\begin{equation}\label{c6eq:wrench_constr}
A_{cl}f_l \leq b_{cl} \quad \forall t: t \leq t_f.
\end{equation}
Notice that these constraints are consistent only if the foot is in contact. Thus, they have not to be applied on the swing limb, as the corresponding required wrench should be null when $t<t_{impact}$. The constraints applied on the swing foot attempt to catch the hybrid nature of the system by varying in time. 
Formally:
\begin{equation}\label{c6eq:fr_cont_costr}
\begin{cases}
	f_{r} = 0_6 &\forall t:\; t < t_{impact}\\
	A_{cr}f_r \leq b_{cr}  &\forall t:\; t_{impact} \leq t \leq t_f.
\end{cases}
\end{equation}
By considering the CoM dynamics only, we elude the complications introduced by time-varying constraints on the robot configuration after the establishment of a new contact (e.g. the velocity of the corresponding foot should be zero).
\subsection{Cost function definition}\label{c6sec:cost}

The cost function ${\Gamma}$ possesses different components that act only before or after the impact.
Formally, it has the following expression:
\begin{IEEEeqnarray}{rCl}
\IEEEyesnumber \phantomsection \label{c6eq:cost}
\Gamma =& \frac{1}{2} & \left(  \int_{0}^{t_f} \norm{ \gamma(\tau) - \gamma^d(\tau)}^2_{K_\gamma}  \dif\tau + \right. \IEEEyessubnumber \label{c6eq:cost_gamma}\\
 &+&\int_{\bar{t}_{imp}}^{t_f} \norm{\gamma(\tau) - \gamma^d(\tau)}^2_{K_\gamma^{imp}}  \dif\tau + \IEEEyessubnumber \label{c6eq:cost_gamma2}\\
 &+&\int_{0}^{t_f}  \norm{f(\tau)}^2_{K_f} \dif\tau +\IEEEyessubnumber \label{c6eq:cost_f}\\
 &+& \norm{\gamma(t_f) - \gamma^d(t_f)}^2_{K_\gamma^{imp}}\Bigg). \IEEEyessubnumber \label{c6eq:cost_gamma_ter}
\end{IEEEeqnarray}
$K_{(\cdot)}^{(\cdot)}$ is a real positive semi-definite matrix of gains with suitable dimensions, accounting also for unit of measurement mismatches inside ${\Gamma}$. $\bar{t}_{imp}$ is the minimum between $t_{impact}$ and $t_f$ (to avoid negative integrals when the impact is expected to occur after $t_f$) while, for the sake of simplicity, the initial time instant is set to zero.
Note that it is possible to vary the cost applied to the state $\gamma$ after the impact through the use of the matrix $K_\gamma^{imp}$. 
Finally, a terminal cost term, weighted by the same matrix $K_\gamma^{imp}$, models the finiteness of the control horizon.

Particular attention has been payed to the choice of the references, here expressed with the superscript $d$. Indeed, even if the model carries no information on how the step is made, the CoM dynamics highly influences the resultant motion.
The choice of the references are therefore crucial for performing a step correctly, since they are the only insight we could give to the controller about the desired posture of the robot. On the other hand, to provide a precise CoM trajectory would reduce the benefits and the efficiency of the MPC controller.
After the completion of the step, we would like to have the CoM over the convex hull centroid. Nevertheless, this objective does not constraint the CoM trajectory during the step. Ideally, we could leave the optimizer to come out with an optimal trajectory, given initial and final conditions. For this reason, we decide to weight the traverse components (i.e. $x$ and $y$) of the CoM position only in the terminal cost Eq.\eqref{c6eq:cost_gamma_ter} and after the step is made, Eq.\eqref{c6eq:cost_gamma2}. On the other hand, the $z-$component of the CoM position is continuously weighted, allowing to have some authority over the CoM trajectory, useful to perform the step movement correctly. The CoM velocity and the angular momentum are always minimized.

As a last term, we weight the requested reaction forces with the goal of avoiding impulsive responses. Additionally, having smooth references for the desired wrenches allows to reduce the tracking error introduced by the underlying momentum controller.

\subsection{References definition for the whole-body torque controller}
The outputs of the MPC strategy are used as references for the whole-body torque controller described in Section \ref{sec:momentum}.
In particular, $f^{ref} = f$, i.e. the contact wrenches are directly used as references.
Regarding the desired joint positions $q_j^{ref}$, we solve an inverse kinematics problem, viz., we impose a desired pose for the swing foot, following as close as possible the center of mass trajectory defined by the MPC controller.


\section{MPC as a Step Trigger}\label{trigger}
Let us consider the scenario where the robot has to perform a step because of an external perturbation. In principle we could leverage the presented MPC formulation to define the moment in which to start the motion, up to now considered as a datum. This moment can be defined as the instant where the controller is not able to bring the CoM back to a desired equilibrium point, given the present support configuration. In other words, the constraints related to the balancing configuration prevent the controller to recover from the push. Thus, it is necessary to step and change balancing configuration in order to avoid a fall. The availability of a prediction horizon particularly suits this idea. Hypothetically, if $t_f = \infty$, as soon as the robot is pushed, we could predict whether or not the controller will be able to absorb the disturbance completely. In practice this is not true. The finiteness of the prediction horizon hides the actual recovery capabilities of the controller, thus it is still necessary to define an heuristic.
By considering the very last predicted state, we can set the following condition:
\begin{equation}\label{eq:step_condition}
\|x_{CoM}(t_f) - x_{CoM}^d\| {+} k_v \|\dot{x}_{CoM}(t_f) - \dot{x}_{CoM}^d\| < \bar{d}.
\end{equation} 
When Eq.\eqref{eq:step_condition} is violated, the robot performs the step. Both $k_v$ and $\bar{d} \in \amsmathbb{R}$ are user-defined parameter affecting the sensitivity to pushes. The smaller $\bar{d}$ the more the robot will be inclined to take a step, while $k_v$ allows to regulate the relative importance of the two errors. 
Notice that Eq.\eqref{eq:step_condition} does not depend directly on the feet dimensions. 

In order to be effective, this heuristic needs the MPC strategy to be in charge of sending references to the robot, even if the step will not be performed at all. Indeed, Eq.\eqref{eq:step_condition} is based on a prediction which assumes that all the computed wrenches are directly applied on the system. Nevertheless, notice that here the goal is different from what has been presented in Sec. \ref{c6sec:cost}. In a sense, the robot should do its best to avoid a step. As a consequence, looking at Eq.\eqref{c6eq:cost}, $\bar{t}_{imp}$ will be equal to $t_f$ (i.e. the step will not occur at all), while $K_\gamma$ should be equal to $K_\gamma^{imp}$. In practice, the cost function should resemble the case where the impact already happened, in order to weight correctly the whole state terms.

\section{Quadratic Programming Transcription} \label{sec:qp}

We solve the finite-horizon optimal control problem presented in the previous section by using direct numerical methods. In particular we convert the original optimal control problem into a single Quadratic Programming (QP) problem to be solved at each time step. 

The general form of a QP problem is the following:
\begin{IEEEeqnarray}{CRCL}
	\IEEEyesnumber \phantomsection \label{c6eq:optim}
	\minimize_\chi& \frac{1}{2}\chi^\top H \chi &+& \chi^\top \mathbb{g} \IEEEyessubnumber \label{c6eq:optim_cost}\\ 
	\text{subject to:} & A\chi &\leq& b \IEEEyessubnumber
\end{IEEEeqnarray}
where $\chi \in \amsmathbb{R}^n$ is the set of optimization variables, $H \in \amsmathbb{R}^{n \times n}$ a positive semi-definite matrix (the ``Hessian'' matrix), $\mathbb{g} \in \amsmathbb{R}^n$ the ``gradient'' vector, while $A \in \amsmathbb{R}^{n_g \times n}$ and $b \in \amsmathbb{R}^{n_g}$ define the set of $n_g$ constraints applied to the optimization variables.

\subsection{Model transcription}

We discretize the model using forward Euler formulation. Different approaches may have been chosen, as described in \cite{BettsPractical2010}, but we decided that Euler integration was suitable thanks to its simplicity.
Discretizing Eq.\eqref{c6eq:model}, we obtain:
\begin{equation}\label{c6eq:discrmodel}
\gamma(k+1) = Ev_\gamma \gamma(k) + F_\gamma f(k) +G_\gamma + S_\gamma ^0  
\end{equation}
where
$
Ev_\gamma = 1_9 + \mathrm{d}t\tilde{E}v_\gamma,\, F_\gamma = \mathrm{d}t\tilde{F}_\gamma, \, G_\gamma = \mathrm{d}t\tilde{g}_\gamma, \, S_\gamma ^0 = \mathrm{d}t\tilde{S}^0_\gamma
$
while $k \in \mathbb{N}$, $k = 0,\cdots\, , \, N-1$ is the discrete time, $N=\ceil{t_f/\mathrm{d}t}$ and $\dif t\in \amsmathbb{R}$ is the time step length.

We can now define the optimization vector $\chi$ as the collection of all the states and control variables over the horizon. 
Formally:
\begin{equation}
\chi =
\begin{bmatrix}
\gamma(1)\\
f(0)\\
\vdots\\
\gamma(k)\\
f(k-1)\\
\vdots\\
\gamma(N)\\
f(N-1)
\end{bmatrix}, \: \chi \in \amsmathbb{R}^{9N+12N}.
\end{equation}
We thus opted for leaving the state $\gamma$ as an optimization variable. This choice allows for an easy reformulation of the cost function, while preserving a particular sparse structure which may be useful for the optimizer.

According to the receding horizon principle, at each time step, $f(0)$ will be applied to the system, through a modified version of the balancing controller implemented on iCub, as presented in Sec. \ref{sec:model}. 

A crucial point to be considered, is the definition of the time instant at which the impact occurs.
We assume the impact to be impulsive and taking place at the beginning of a time step which we denote with $k_{impact}$.
We further assume that the control input $f(k)$ is applied piecewise constantly, i.e. constant from instant $k$ to $k+1$. 
Note that, in view of the above two assumptions, setting $k_{impact} = 0$ implies that both feet are at contact already at the initial time $k = 0$.

As mentioned previously, we do not directly model the distance between the swing foot and the ground, but we compute the position and the expected impact time at the beginning of the push strategy.
The following procedure describes how we update the expected impact instance throughout the proposed MPC controller.
Assuming the initial time instant to be always set to zero, we start with a value of $k_{impact}$ equal to $\lceil t_{impact}/\mathrm{d}t \rceil$. At each controller execution, we set the new value to $\max \{k_{impact} -1, ~ 1\} $, which implies that, if the impact has not occurred yet, we saturate $k_{impact}$ to $1$, i.e. we expect the impact to occur at the second time step in order to avoid requiring wrenches on a swinging limb.
If the impact has occurred, instead, $k_{impact}$ is correctly set to $0$.
Note that the formal definition of the impact time and the trajectory control of mechanical system with unilateral constraints (as it is our case), are still an open research problem \cite{rijnen2015optimal,biemond2013tracking}. 


It is now possible to write the discretized dynamics as an equality constraints, transforming the initial continuous time optimal control problem into Eq.\eqref{c6eq:optim}. By expanding Eq.\eqref{c6eq:discrmodel} over the horizon we get the following constraint:
\begin{equation} \label{c6eq:qp_ev}
\left(Ev - e_\gamma\right) \chi = -G - Ev_{\gamma_0} - S^0
\end{equation}
where $Ev \in \amsmathbb{R}^{9N\times 21N}$ and $G, \:Ev_{\gamma_0}, \:S^0 \in \amsmathbb{R}^{9N\times 1}$. The matrix $e_\gamma\in\amsmathbb{R}^{9N\times 21N}$ selects the terms related to state $\gamma$ from vector $\chi$, while $Ev$ has a peculiar banded structure due to the choice of vector $\chi$. 
In particular, different repetitions of $Ev_\gamma$ and $F_\gamma$ are included, replicating Eq.\eqref{c6eq:discrmodel} for each time step in the control horizon.

The dependency from $\gamma(0)$ is considered through matrix $Ev_{\gamma_0}$, constituted by only zeros, except from the first block of nine rows, which are equal to $Ev_\gamma$.
Finally $G$ considers the gravitational effects, while $S^0$ contains the constant terms related to the angular momentum.

A similar approach can be used to transcribe the constraints presented in \ref{c6sec:constraints}:
 \begin{equation}\label{c6eq:f_discr}
 A_{hor}\: \chi \leq b_{hor}.
 \end{equation}
 Matrices $A_{hor}$ and $b_{hor}$ condense the constraints of Eq.\eqref{c6eq:wrench_constr} and Eq.\eqref{c6eq:fr_cont_costr} for each time instant. In order to avoid a varying number of constraints from one MPC execution to the other, we adopted a simple expedient. An upper and lower bound on the normal force $\mathbb{f}_z$ are added, among friction and CoP constraints. In order to set the wrench on the swing foot to zero, it is simply necessary to set both the previously mentioned upper and lower bound to zero. All the other components of the wrench are automatically set to zero thanks to the friction and CoP constraints. Thus, it is simply necessary to correctly handle the terms inside $b_{hor}$ without changing the number of constraints. 
\subsection{Cost function transcription}
The last stage of the transcription process involves the cost function of Eq.\eqref{c6eq:cost}. 
As before, this process consists in a discretization stage followed by a reformulation phase necessary to express the cost function in terms of $\chi$. This latter stage is omitted here, but it simply exploits the appropriate \emph{extractor} matrices, as it has been done for the model transcription through $e_\gamma$.

Starting from the first term, Eq.\eqref{c6eq:cost_gamma}, the discretization corresponds to 
\begin{IEEEeqnarray}{C}
\IEEEyesnumber \phantomsection
\eqref{c6eq:cost_gamma} \Rightarrow \Gamma_\gamma = \frac{1}{2}\sum_{k=1}^{N} \norm {\gamma(k) - \gamma^d(k)}^2_{K_\gamma} 
\end{IEEEeqnarray}
noticing that the cost is evaluated starting from 1, avoiding to consider the feedback $\gamma(0)$.

Similarly, the other contributions can be discretized as:
\begin{IEEEeqnarray}{RCL}
\eqref{c6eq:cost_gamma2}{+}\eqref{c6eq:cost_gamma_ter} &\Rightarrow&  \Gamma_\gamma^{imp} = \nonumber\\
	 &=&\frac{1}{2}\sum_{k=\bar{k}_{imp}}^{N} \norm{\gamma(k) - \gamma^d(k)}^2_{K_\gamma^{imp}},
 \IEEEyessubnumber\\
\eqref{c6eq:cost_f} &\Rightarrow& \Gamma_f = \frac{1}{2}\sum_{k=0}^{N-1} \norm {f(k)}^2_{K_f} \IEEEyessubnumber,
\end{IEEEeqnarray}
where $\bar{k}_{imp}$ is the minimum between $N$ and $k_{impact}$, so that $\Gamma_\gamma^{imp}$ is composed by at least one term, as if there was a terminal cost like in ${\Gamma}$. 

We can exploit the simple structure of the cost function to further weight the difference of desired reaction forces between two subsequent time-steps, in order to avoid abrupt changes in the control input. 
This new cost term in the discrete case has the following form:
\begin{equation}
\Gamma_{df} = \frac{1}{2}\sum_{k=0}^{N-1} \norm {f(k)-f(k-1)}^2_{K_{df}} 
\end{equation}
with $K_{df}$ a suitable positive semi-definite matrix of gains and initialized with $f(-1)$, considered as a datum, e.g. at the beginning it corresponds to the wrench necessary to satisfy the equilibrium of the robot momentum.

Finally, the overall cost function $\Gamma$ is:
\begin{equation}
\Gamma = \Gamma_\gamma + \Gamma_\gamma^{imp} + \Gamma_f +\Gamma_{df}.
\end{equation}

\section{Simulation Results} \label{sec:simulation}

\begin{figure*}[t]
	\centering
    \subfloat[Side step] {\includegraphics[width=.33\textwidth]{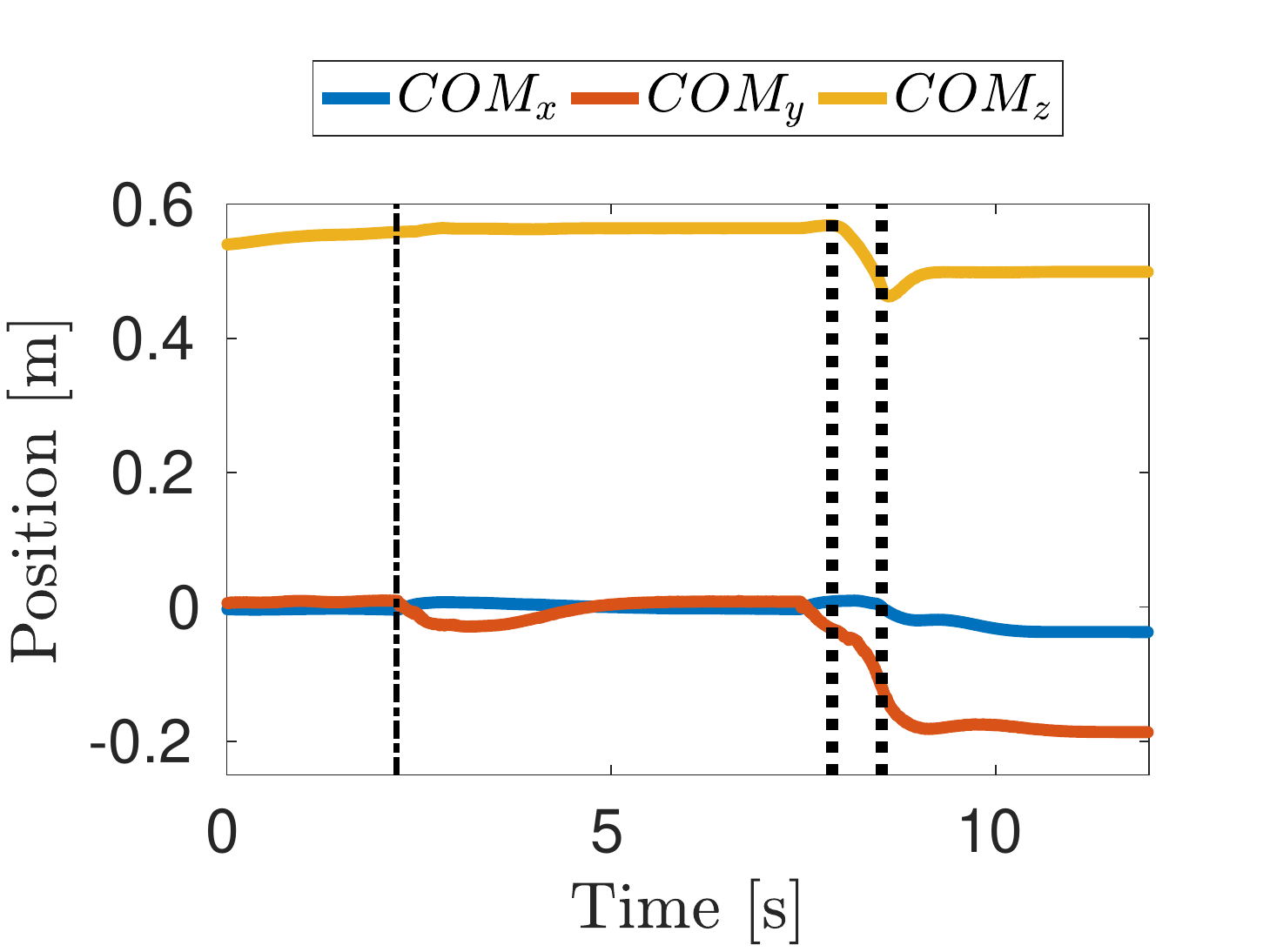}\label{fig:side}}
	\subfloat[Back step] {\includegraphics[width=.33\textwidth]{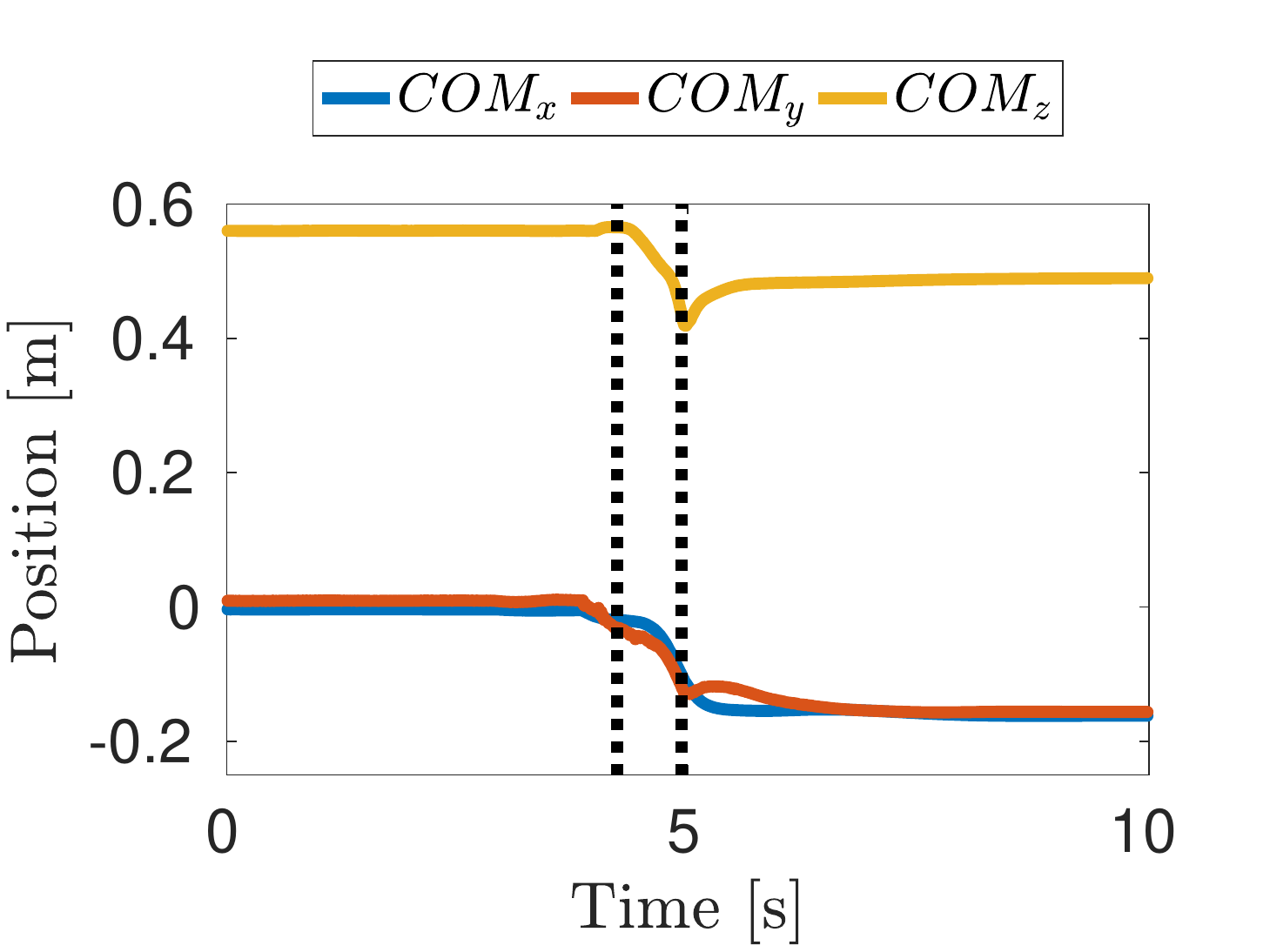}\label{fig:back}}
	\subfloat[Forward step] {\includegraphics[width=.33\textwidth]{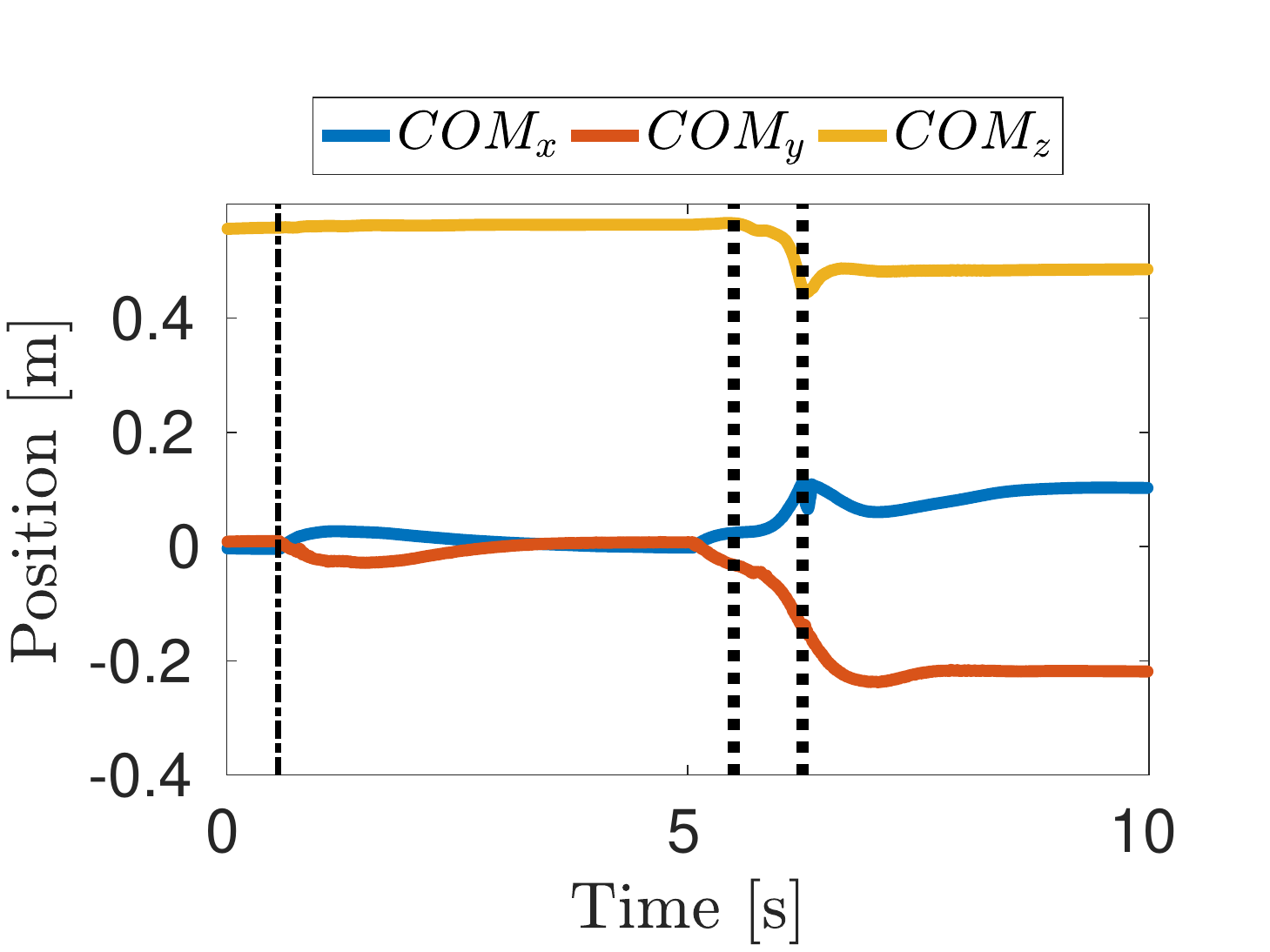}\label{fig:front}}
    \caption{CoM evolution during three different experiments of push recovery. The single dashed lines in \protect\subref{fig:side} and \protect\subref{fig:front} show a not-enough-strong push to violate the condition in Eq. \eqref{eq:step_condition}. The external push force has been applied, with respect to the lateral axis, at $20\degree$, $-20\degree$ and $45\degree$ for the cases \protect\subref{fig:side}, \protect\subref{fig:back} and \protect\subref{fig:front} respectively.}
	\label{fig:exp_com}
\end{figure*}

%
%

\begin{figure}[t]
	\centering
\includegraphics[width=.9\columnwidth]{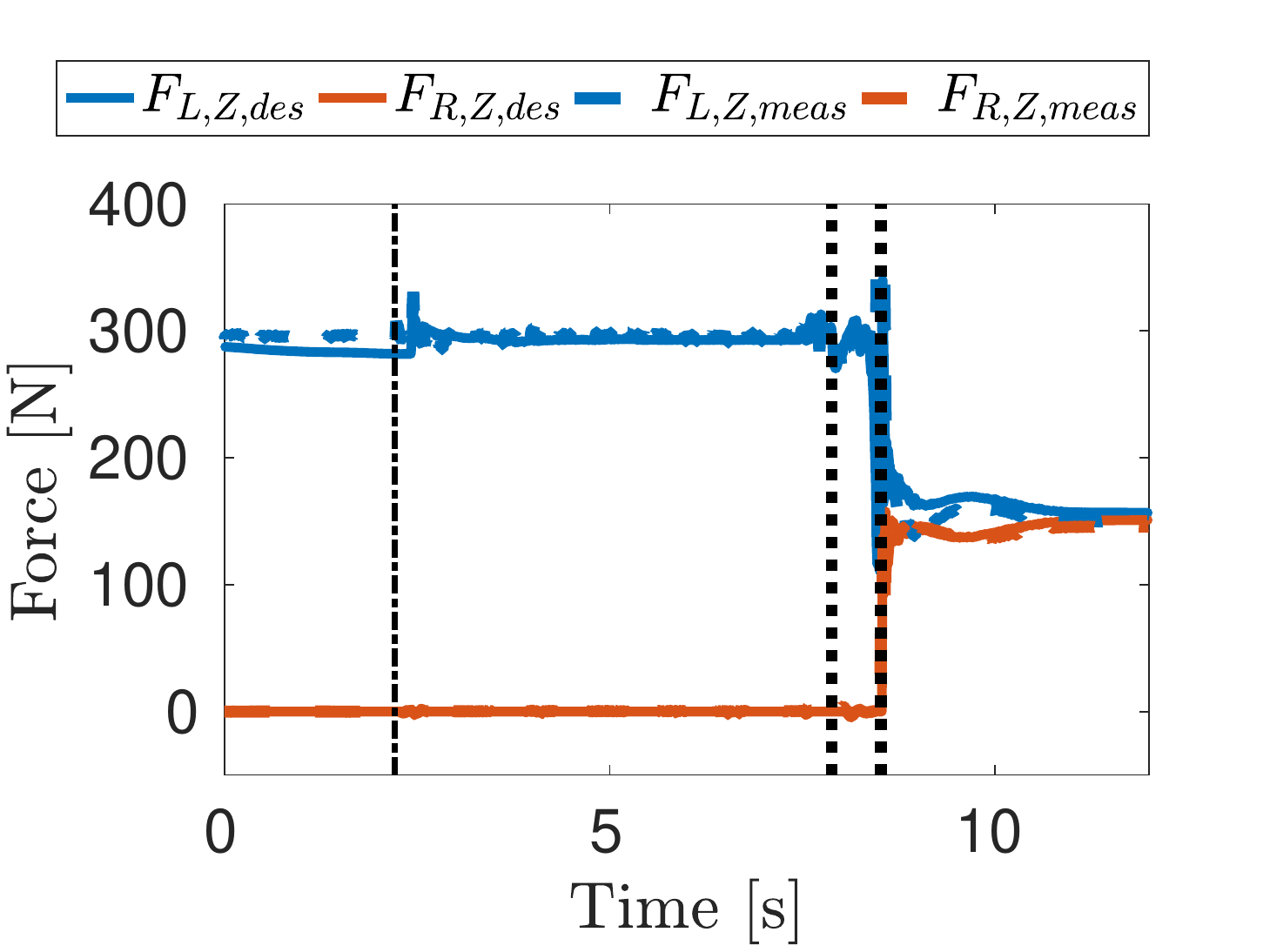}
   \caption{Vertical components of the contact forces during the side-step experiment. Measured forces are plotted with dashed lines, while desired forces with the solid lines.} \label{fig:force}
\end{figure}

\begin{figure}[t]
	\centering
	\includegraphics[width=.9\columnwidth]{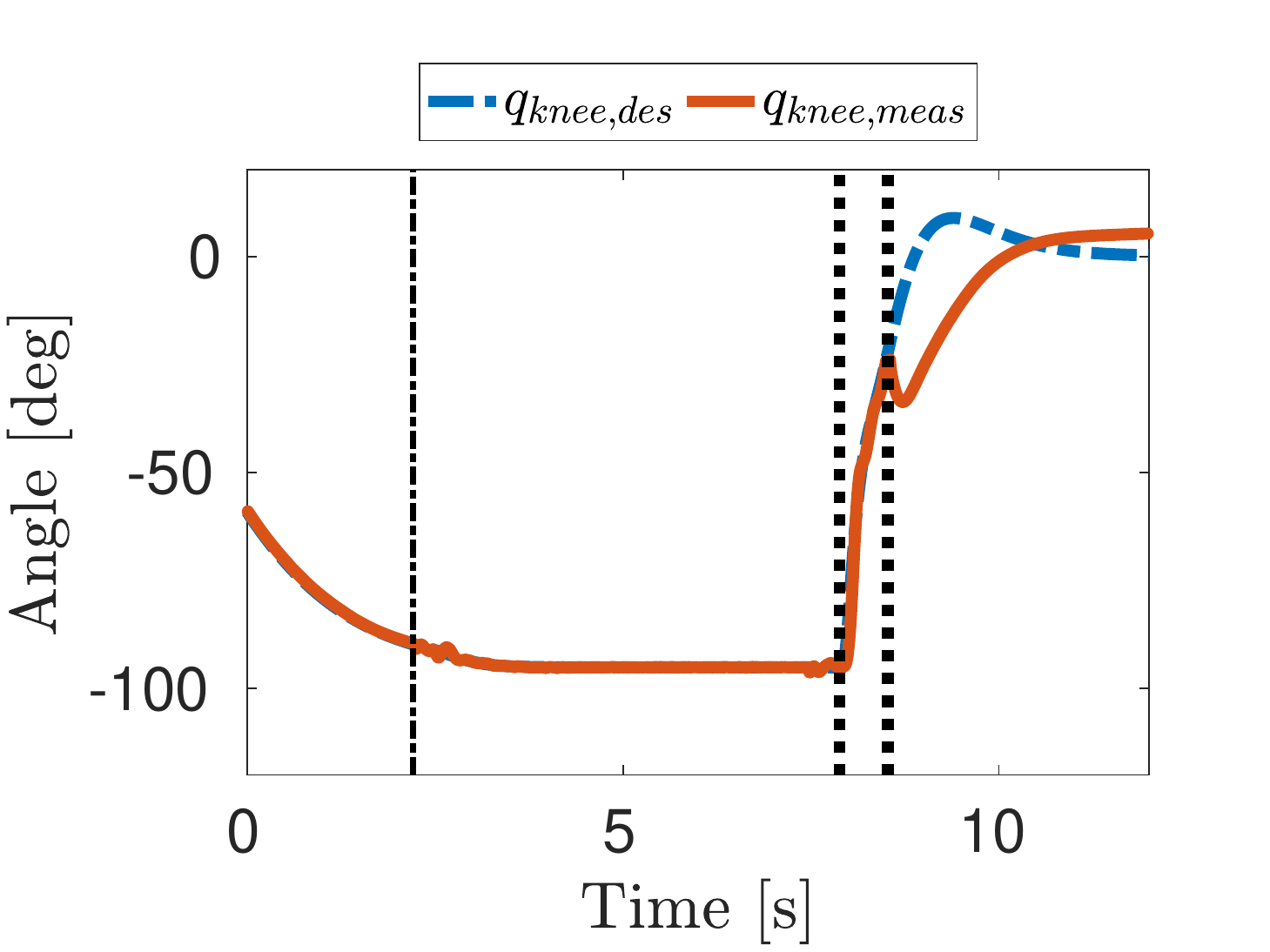}
   \caption{Position tracking error of the right knee joint. 
   The knee absorbs the impact with the ground, as it can be noticed by the peak in the error right after the impact.}\label{fig:knee}
\end{figure}


The presented MPC approach has been tested in the Gazebo simulator \cite{Koenig04} by using the iCub model.
The iCub humanoid robot \cite{metta2005robotcub} has been conceived to study developmental cognitive systems. It possesses 53 actuated joints, but only a total of 23 degrees of freedom (DoF) are used for balancing tasks, i.e. we do not consider those located in the eyes and in the hands.

Driven by the need of fast prototyping, the presented controller has been developed using the MATLAB\textsuperscript{\textregistered}/Simulink\textsuperscript{\textregistered}  environment, taking advantage of  WBToolbox library \cite{RomanoWBI17} and YARP Plugins \cite{YarpGazebo2014} to establish connection with the simulator.
In order to test the presented MPC scheme, we conceived a simple stepping scenario, where the robot, balancing on its left foot, will use the right foot to take a step.
This simple scenario allows us to test the performance of the proposed controller with a single contact activation. 

We present the results of different pushes, applied on the traverse plane, with an angle with respect to the lateral axis (pointing to the right of the robot), of $20\degree$ (Fig. \ref{fig:side}), $-20\degree$ (Fig. \ref{fig:back}) and $45\degree$ (Fig. \ref{fig:front}).

We choose a time step of $10\mathrm{ms}$, coincident with the rate of the whole-body torque controller, and a controller horizon of $N=25$.
We noticed that the chosen value of $N$ is sufficient to allow the effectiveness of the strategy when the robot is pushed from different directions. 
The push is nearly impulsive, applied on the chest with a magnitude around $100\mathrm{N}$, which is about one third of the robot weight force.

Figure \ref{fig:exp_com} shows the CoM evolution for the three experiments, i.e. with different directions of pushes.
It can be noticed in both Figures \ref{fig:side} and \ref{fig:front} that two pushes occur. The first one does not violate the condition in Eq.\eqref{eq:step_condition}, thus it does not force the robot to take a step.
The second one, instead, triggers a change in the support feet configuration, and as a consequence, a new desired configuration for the CoM.

Figure \ref{fig:force} represents the tracking of the desired vertical forces output by the MPC controller for the side-push experiment.
Remarkably, the normal force on the right foot appears to be tracked also across the step. 
Figure \ref{fig:knee} shows one of the benefits of torque control. 
The tracking of the joint position reference on the right knee undergoes a strong perturbation after the step. When hitting the ground, the intrinsic compliance introduced by torque control allows to absorb the impact, especially on this joint. This induce a peak of $30\degree$ of tracking error, but the robot is still able to balance. 
In addition, torque control helps avoiding problems related to a not perfect placement of the swing foot before the impact.


Summarizing, the presented controller allows the robot to recover from pushes of various intensity and directions, while remaining able to perform involved step movements. 

\section{Conclusions and Future Work} \label{sec:conclusion}
The proposed controller adopts a slightly approximated model of the robot linear and angular momentum dynamics in a predictive framework. It allows to take into account step movements by varying the structure of constraints and cost functions across the change of contacts configuration. The uncertainty on its actual time instant is considered by a ``shift" in the prediction window. An heuristic is also employed to determine a condition for stepping. The contact wrenches are assumed to be control inputs and realized by the robot through a modified version of the iCub momentum-based whole-body torque controller. 

Taking \cite{Dafarra2016} into a comparison, this approach avoids the definition of a CoM trajectory along the step, leaving the responsibility to the optimizer rather than to the designer. As a return, the proposed strategy presents higher robustness properties with the drawback of an increased computational complexity.
Since this approach has been meant to provide reaction in real-time, efforts have to be payed in order to reduce the time needed to get a solution. At the present time it takes almost $0.1$ seconds to solve an instance of the presented formulation on a machine running Ubuntu 16.04. The PC is equipped with a quad-core Intel\textsuperscript{\textregistered} Core i5@2.30GHz and 16GB of RAM. MOSEK\textsuperscript{\textregistered} is the selected solver, accessed through the MATLAB\textsuperscript{\textregistered} interface \texttt{CVX} \cite{cvx}. An optimized C++ version of this strategy is currently under development. In addition, for a practical implementation on the real robot, the delay between the feedback and the actual generation of references should be reduced as much as possible. Alternatively, we might consider a one step delay for the application of the control action, directly in the model.

An additional entry on the checklist of future improvements consists in providing the MPC controller with information about the leg kinematics. This would allow to insert the step peculiar characteristics (step duration and location) directly inside the optimization.

\addtolength{\textheight}{0cm}   

\addcontentsline{toc}{section}{References}

\bibliography{IEEEabrv,Bibliography}

\end{document}